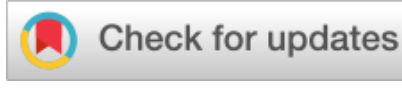

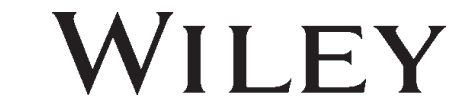

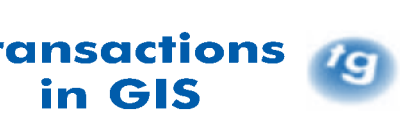

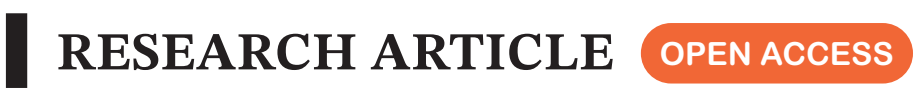

# Cross-Modal Urban Sensing: Evaluating Sound–Vision Alignment Across Street-Level and Aerial Imagery

Pengyu Chen[1] | Xiao Huang[2] | Teng Fei[3,4] | Sicheng Wang[1]

[1]Department of Geography, University of South Carolina, Columbia, South Carolina, USA | [2]Department of Environmental Sciences, Emory University, Atlanta, Georgia, USA | [3]University of Canterbury, Christchurch, New Zealand | [4]School of Resource and Environmental Sciences, Wuhan University, Wuhan, China

**Correspondence:** Sicheng Wang (sichengw@mailbox.sc.edu)

## ABSTRACT

Environmental soundscapes carry rich ecological and social information, yet remain underutilized in geographic analysis. This study introduces a unified cross-modal evaluation framework to examine how urban sounds align with visual representations from street-level and aerial perspectives, and how different visual representation strategies influence this alignment. We integrate geo-referenced sound recordings from London, New York, and Tokyo with corresponding street-view and remote-sensing imagery, applying embedding-based methods (AST for audio; CLIP and RemoteCLIP for imagery) and segmentation-based methods (CLIPSeg and Seg-Earth OV). Results show that embedding-based models capture stronger semantic alignment between sound and imagery, particularly for street-level views, while segmentation-based models, especially from aerial imagery, more effectively reveal interpretable ecological patterns when mapped to Biophony, Geophony, and Anthrophony (BGA) categories. These findings highlight the complementary strengths of embeddings for fine-grained semantic matching and segmentation for ecological interpretation, offering a reproducible evaluation framework for incorporating sound into multimodal urban sensing.

## 1 | Introduction

Urban environments are complex, multilayered systems shaped by interacting social, ecological, and physical processes. In recent years, the integration of multimodal data, particularly visual data such as aerial imagery and street-level photographs, has become central to urban analysis and monitoring (Zhang, Salazar-Miranda, et al. 2024; Heidler et al. 2023; Biljecki and Ito 2021). However, one sensory modality remains conspicuously underexplored: sound. As the acoustic signature of place, urban soundscapes encode rich information about human activity, ecological presence, and environmental conditions (Zhao et al. 2023; Arzberger et al. 2025; Kim et al. 2025). Prior research has demonstrated their relevance to topics such as public health (Aletta et al. 2025; Kong and Han 2024), biodiversity (Xu et al. 2023), and socio-environmental equity (Chen et al. 2021), yet systematic approaches to incorporating sound into spatial frameworks are still rare.

Recent advances in multimodal machine learning have opened new avenues for bridging the divide between auditory and visual modalities in urban sensing (Guzhov et al. 2022; Girdhar et al. 2023; Heidler et al. 2023). Cross-modal similarity analysis, quantifying how closely sounds correspond to visual features of their surroundings, provides a way to integrate these complementary forms of environmental information. While visual data captures the structural and material composition of space, sound encodes dynamic, temporal, and often hidden processes such as movement, density of activity, and ecological presence. Understanding how these modalities align is therefore essential for building more complete representations of urban environments, connecting what cities look like with what they sound like.

Despite the growing availability of multimodal models, little work has examined how different forms of visual representation influence this alignment. Embedding-based approaches capture semantic and perceptual similarity within a shared feature space, whereas segmentation-based methods provide explicit compositional and ecological information about land-cover elements. Comparing these two strategies offers an opportunity to assess both the strength and the interpretability of sound–vision correspondence. Embedding models can reveal abstract perceptual relationships, while segmentation models can expose tangible ecological structures that shape soundscapes. This complementary perspective defines the core rationale of our study.

To advance this line of inquiry, our research addresses two central questions:

1. To what extent do environmental soundscapes correspond with visual representations of urban space, spanning both street-level imagery and overhead remote-sensing views?
2. How does the choice of visual representation strategy—embedding-based or segmentation-based—affect the quality and interpretability of this cross-modal alignment?

To explore these questions, we compare two fundamentally different approaches to visual representation: (1) embedding-based models, such as CLIP (Radford et al. 2021) and RemoteCLIP (Liu et al. 2024), which encode visual scenes into high-dimensional latent feature spaces; and (2) semantic segmentation models, including CLIPSeg (Lüddecke and Ecker 2022) and SegEarth-OV (Li, Liu, et al. 2024), which provide interpretable, class-level descriptions of scene content. These visual features are paired with audio embeddings extracted from the Audio Spectrogram Transformer (AST) (Gong et al. 2021b), a state-of-the-art model for acoustic scene understanding. The alignment between modalities is then quantified through two distinct pathways. In the embedding-based approach, we calculate the cosine similarity between the audio and visual feature vectors to measure semantic correspondence in a shared latent space. In the segmentation-based approach, we derive the proportion of different visual classes from imagery (e.g., vegetation, water, buildings) and map these proportions to ecological sound categories: Biophony, Geophony, and Anthrophony (BGA), to create an interpretable ecological profile for comparison. By contrasting these two methods, we evaluate not only the strength of sound–vision alignment but also the interpretive value each representation strategy offers.

This study offers two key contributions:

- It introduces a unified cross-modal evaluation framework for urban sensing that integrates geo-referenced soundscapes with both street-level and aerial imagery, enabling multi-scale analysis of visual–acoustic correspondence.
- It conducts a systematic comparison between embedding-based and segmentation-based visual representations, assessing their respective capacities to capture and interpret cross-modal alignment between urban sounds and scenes.

Overall, this work contributes to the emerging field of multimodal urban sensing by offering foundational insights into how acoustic information can be meaningfully incorporated into geospatial analysis. By examining the interplay between the sights and sounds of urban life, we aim to support more comprehensive, ecologically grounded, and socially attuned understandings of cities.

## 2 | Related Work

### 2.1 | Urban Environmental Soundscape Analysis

Urban soundscape research has increasingly evolved from treating sound as a nuisance or byproduct of urbanization to understanding it as a meaningful dimension of socio-ecological systems (Aletta et al. 2025; Xu et al. 2023). Soundscapes are now recognized as vital indicators of environmental quality, biodiversity, public health, and psychological well-being (Chen et al. 2021; Arzberger et al. 2025; Kong and Han 2024).

Broadly, the literature reflects three main research directions. The first focuses on sound perception and human experience, using controlled experiments or immersive simulations to assess how specific sound types—for example, birdsong, human chatter, or traffic—shape perceptions of comfort, sociability, and appropriateness in urban environments (Jingwen Cao 2024; Huang et al. 2024). Natural sounds tend to promote tranquility, while human-generated sounds are associated with increased sociability.

The second strand centers on automated sensing and ecological inference, emphasizing scalable, sensor-driven monitoring frameworks such as Smart Soundscape Sensing (SSS) (Wang et al. 2023). These approaches combine sensor networks with machine learning to infer ecological patterns, such as avian biodiversity, from acoustic signals (Arzberger et al. 2025), offering an efficient alternative to manual field surveys.

The third research direction incorporates contextual and text-driven analytics, leveraging ancillary metadata—such as emotional tags, time of day, or stated visit purposes—to classify and map soundscapes across urban contexts (Kim et al. 2025). Participatory methods, such as causal loop diagrams, have also been introduced to capture the complex feedback loops between acoustic environments, social equity, and urban morphology (Aletta et al. 2025).

These three strands have significantly advanced our understanding of environmental soundscapes, although they remain largely sound-centric and single-modality. In parallel, numerous multimodal models have been introduced since the development of WaveNet (van den Oord et al. 2016), yet few studies have systematically examined the bidirectional relationship between sound and vision in urban contexts or assessed how different visual representation strategies influence our ability to model this relationship. Our study addresses this gap by explicitly comparing embedding-based and segmentation-based visual features in relation to urban soundscapes, aiming to promote a more balanced, interpretable, and multimodal analysis of place.

## 2.2 | Visual Representations of Urban Environments

Visual data have long served as a cornerstone in urban research, offering scalable and tangible representations of the built environment. With the advent of large-scale datasets—ranging from street-level panoramas to high-resolution aerial imagery—and the rise of deep learning, visual analytics have significantly expanded their role in characterizing urban form and function (Zhang, Salazar-Miranda, et al. 2024; Yang et al. 2024). Attributes such as greenery, walkability, and perceived safety are now routinely quantified using semantic segmentation, object detection, and scene classification techniques (Zhang, Li, and Zhang 2024; Ito et al. 2024). In parallel, aerial imagery provides macro-scale insights into land cover distribution, urban expansion, and environmental risk (Boulila et al. 2021; Pešek et al. 2024; Ray et al. 2024), supported by public datasets like LoveDA (Wang et al. 2021), FLAIR (Garioud et al. 2022), and DOTA (Ding et al. 2021).

These advances are powered by an ecosystem of powerful visual models. Transformer-based architectures such as DETR (Carion et al. 2020), DINOv2 (Oquab et al. 2024), and CLIPSeg (Lüddecke and Ecker 2022) enable robust scene parsing and semantic extraction from ground-level images. For aerial views, models like SegEarth-OV (Li, Liu, et al. 2024), MoCaE (Oksuz et al. 2024), and LSKNet (Li, Li, et al. 2024) continue to improve the accuracy and scalability of land cover segmentation. These methods support scene-level feature analysis—from calculating vegetation ratios to estimating built complexity—and contribute to understanding urban mobility, perception, and planning.

Notably, recent studies have begun to apply visual data to predict acoustic properties. For example, Zhao et al. (2023) used street view images to estimate 15 soundscape indicators, while others have predicted dominant sound types from urban visuals in cities like Fuzhou (Rui et al. 2024). While these studies demonstrate the promise of vision-based acoustic inference, they largely adopt a unidirectional paradigm: treating sound as a target variable to be predicted from static imagery. This approach has several limitations. First, it reduces sound to a label, discarding its temporal richness and contextual specificity. Second, it neglects the synergistic relationship between sight and sound, where each modality can disambiguate or contextualize the other. Lastly, it risks building biased or oversimplified models of urban environments—for example, assuming green equals quiet—overlooking instances where visual cues diverge from acoustic reality. Instead of treating sound as a dependent outcome, we argue that it should be modeled as an independent and co-informative modality. Our study adopts this perspective by directly assessing the semantic alignment between urban soundscapes and visual representations, allowing for a more balanced, multimodal interpretation of place.

## 2.3 | Sound–Visual Embedding Models for Urban Sensing

Recent progress in multimodal learning has enabled sound and vision to be represented within unified latent spaces, opening new opportunities for urban sensing. Instead of treating audio as an auxiliary label, these models regard it as an informative modality complementary to vision, capable of enriching spatial and ecological interpretation.

Early architectures such as SoundNet (Aytar et al. 2016) and L3-Net (Lu et al. 2019) demonstrated that large collections of unlabeled videos could be used to learn audio–visual correspondences in a self-supervised manner. These pioneering efforts revealed that ambient audio encodes meaningful visual cues and vice versa, providing the foundation for cross-modal representation learning. However, the learned associations are coarse and weakly aligned with geographic context, often failing to differentiate acoustically similar urban scenes such as traffic corridors and construction sites.

Subsequent contrastive models, including AudioCLIP (Guzhov et al. 2022), MuLan (Huang et al. 2022), and CLAP (Elizalde et al. 2022), extend the vision–language paradigm to audio, image, and text triplets. By aligning multimodal embeddings through large-scale pretraining, they achieve remarkable zero-shot generalization and capture rich semantic relations between sound and visual content. Yet these models are primarily trained on web videos and consumer media rather than environmental recordings, emphasizing object-level semantics ("car," "dog," "voice") rather than spatial or ecological context. When transferred to urban environments, they may overlook subtler cues—such as vegetation density, surface materials, or reverberation—that shape real soundscapes.

Frameworks such as VATT (Akbari et al. 2021) and ImageBind (Girdhar et al. 2023) represent an important milestone for multimodal alignment. Frameworks like VATT utilize self-supervised learning to align three raw modalities (video, audio, and text), while ImageBind scales this to embed six modalities (audio, image, text, depth, thermal, and IMU) within a single latent space. Their scalability and cross-domain flexibility are significant achievements, but their reliance on self-supervised pretext tasks results in high-dimensional and opaque feature compositions that challenge interpretation. In urban sensing applications, where transparency and spatial reasoning are essential, such models remain difficult to validate or explain.

Complementary approaches leverage co-located sound–image datasets to demonstrate that sound can function as an independent environmental modality rather than a secondary supervision signal. The SoundingEarth dataset (Heidler et al. 2023), for example, pairs ambient audio with aerial imagery to pretrain visual models without manual labels. This framework effectively shows that acoustic data capture spatial and ecological patterns, such as vegetation, water, and built intensity, that are also discernible in remote-sensing imagery. While minor spatial mismatches between recording locations and image footprints may introduce uncertainty,

SoundingEarth provides one of the first systematic evidences that environmental sound can inform visual feature learning at landscape scales.

Taken together, these developments reveal both progress and remaining challenges. Joint-embedding models offer rich semantic correspondence but limited spatial grounding; large-scale self-supervised systems broaden modality coverage yet sacrifice interpretability; and acoustic–visual pretraining datasets demonstrate environmental relevance while still facing data alignment issues. These trade-offs highlight the need for spatially interpretable and ecologically informed approaches, an objective that motivates our comparative analysis of embedding-based and segmentation-based visual representations for urban sound–vision alignment.

## 3 | Dataset

Our study integrates geo-referenced sound and imagery data from three major global cities in Europe, North America, and Asia: London, New York City, and Tokyo. These cities were selected not to achieve statistical representativeness but to include input data with different urban forms and contexts, providing diverse test conditions for sound–vision alignment. London's mixed mid-rise structure and extensive green belts represent a decentralized urban form; Many of the New York samples are drawn from the Central Park area, capturing the juxtaposition of dense urban edges and interior green space; and Tokyo's polycentric morphology combines compact commercial districts with interspersed residential zones. Together, these settings reflect distinct planning traditions, building densities, and spatial organizations across continents, offering a meaningful testbed for evaluating cross-modal sound–vision alignment across heterogeneous urban environments.

In total, 175 sites were sampled, 100 in London, 35 in New York City, and 40 in Tokyo. Corresponding street view and aerial images were collected for each site. Figure 1 displays the spatial distribution of the sampled locations across these cities.

### 3.1 | Environmental Sound Data

Environmental audio recordings were obtained from Radio Aporee (2025), an open platform that hosts crowd-sourced soundscapes annotated with precise geographic coordinates. The dataset consists of short ambient clips reflecting a wide range of urban contexts, including parks, traffic corridors, waterfronts, and residential areas.

To ensure consistency and ecological relevance, we applied several filtering criteria. Recordings dominated by human voices were excluded, as they tend to obscure broader environmental sound patterns. Indoor recordings were removed to maintain alignment with outdoor visual scenes. In addition, clips featuring transient or event-specific sounds—such as alarms, crowd outbursts, or isolated vocalizations—were excluded to focus on stable acoustic environments.

Through this screening process, we compiled a set of audio sample data that reflects the contextual and continuous features of the urban soundscape, making it suitable for comparison with corresponding visual features.

### 3.2 | Visual Data

Two types of imagery were paired with the sound recordings: street-level images, obtained via Google Street View (Google 2025b), and aerial images, sourced from Google Earth (Google 2025a). Street-level imagery provides ground-level perspectives of the environment, capturing details such as built structures, vegetation, and urban texture. Aerial imagery complements this by offering broader spatial context and patterns in land use.

To align visual data with the environmental nature of the sound clips, we excluded indoor views and scenes captured under unusual or transient conditions such as snow, rain, or night-time lighting. These filters ensured that visual representations were stable, interpretable, and relevant to the ambient sound environment.

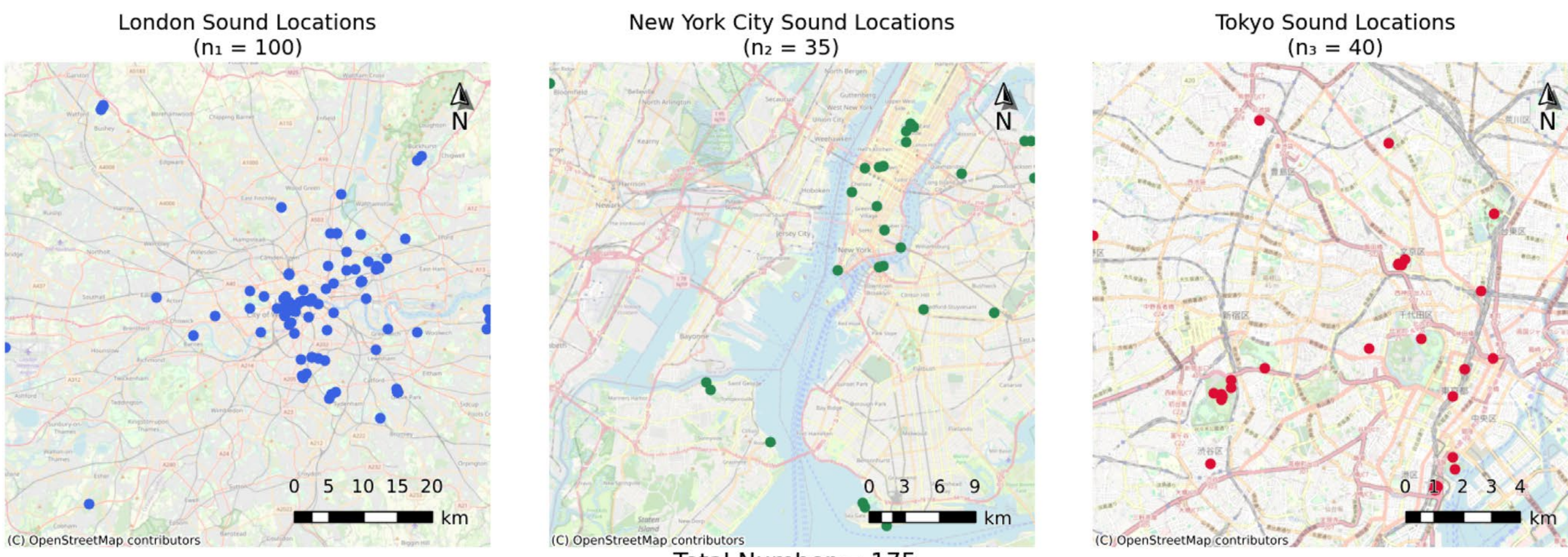

**FIGURE 1** | Geographic distribution of sampled data across London, New York City, and Tokyo.

### 3.3 | Data Alignment and Preprocessing

Each audio clip is associated with a unique geographic coordinate, which serves as the anchor for retrieving both street-level and aerial imagery. All data points are spatially aligned and rigorously screened to minimize contextual inconsistencies. To ensure compatibility with the Audio Spectrogram Transformer (AST) model, each audio recording is segmented into 10-s intervals, following the standard configuration used in the model's pretraining on the AudioSet dataset (Gong et al. 2021b). This duration provides a balanced temporal context for environmental soundscapes—long enough to capture overlapping ambient events but short enough to maintain stable attention and positional embedding behavior in the pretrained AST architecture. Because the majority of recordings represent stationary outdoor ambiences rather than transient sound events, this segmentation does not materially distort their temporal characteristics. This carefully curated multimodal dataset provides a consistent and reliable foundation for subsequent embedding and segmentation-based analyses of urban environments.

## 4 | Methodology

This section delineates the unified cross-modal evaluation framework developed to analyze and compare the alignment between urban soundscapes and visual representations. As illustrated in Figure 2, the pipeline comprises four key phases: data acquisition, feature extraction, similarity computation, and ecological interpretation.

Rather than introducing a new model architecture, this work develops a reproducible evaluation framework that integrates state-of-the-art pretrained models into a coherent, cross-modal workflow. The design enables systematic comparison between two complementary representation strategies: (1) embedding-based alignment, which quantifies abstract perceptual similarity between audio and visual embeddings using high-dimensional encoders (AST for sound; CLIP and RemoteCLIP for imagery); and (2) segmentation-based alignment, which measures structural and ecological correspondence through semantically interpretable land-cover features (CLIPSeg and Seg-Earth OV). The framework further incorporates an ecological mapping layer based on the Biophony–Geophony–Anthrophony (BGA) taxonomy, allowing visual semantics to be interpreted through acoustic–ecological categories. Together, these components establish a consistent evaluation protocol that bridges computer vision, acoustic analysis, and urban ecology, advancing the study of sound–vision correspondence beyond isolated model application.

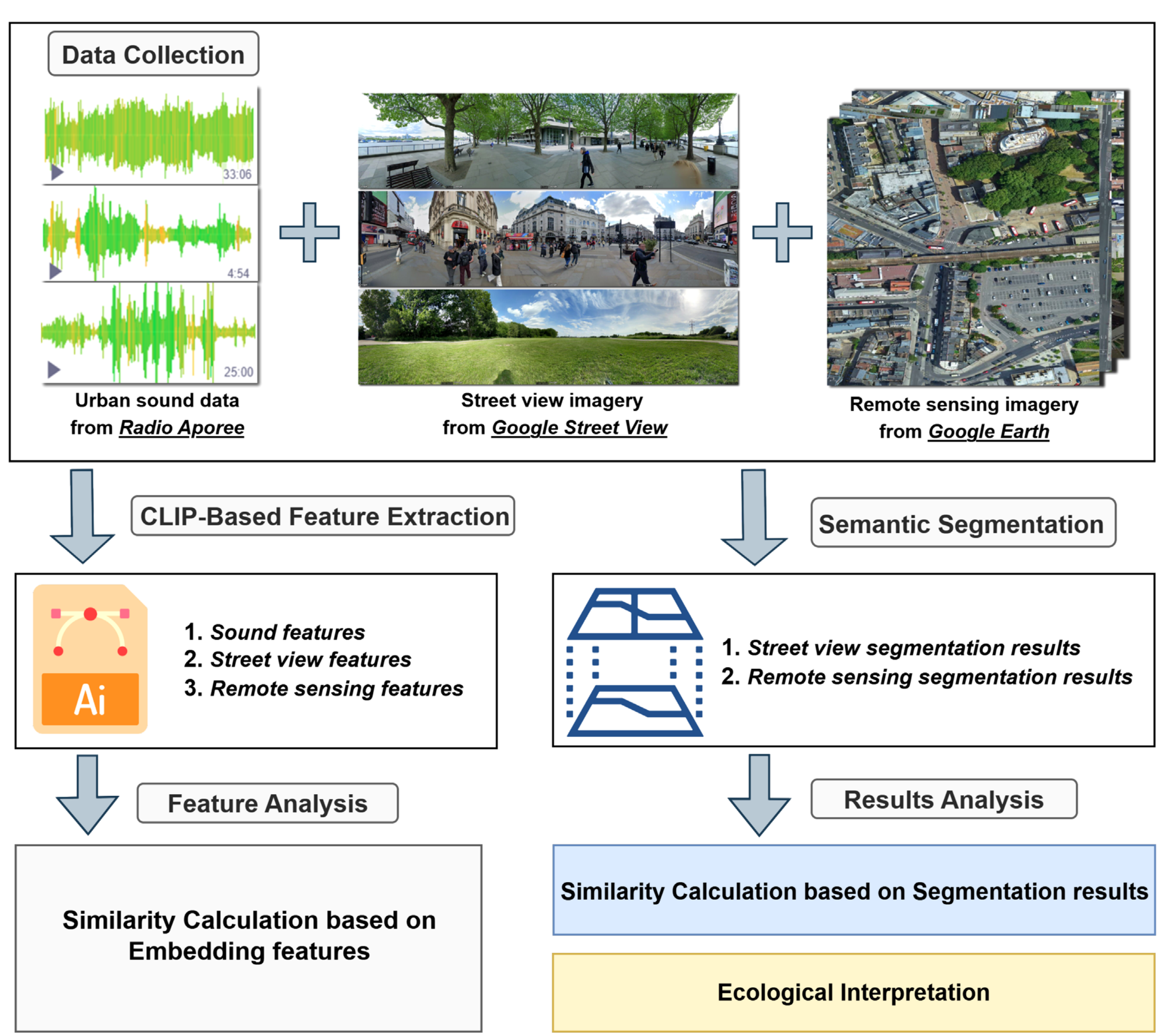

**FIGURE 2** | Overall framework of cross-modal sound–vision alignment, including data collection, feature learning, and similarity computation.

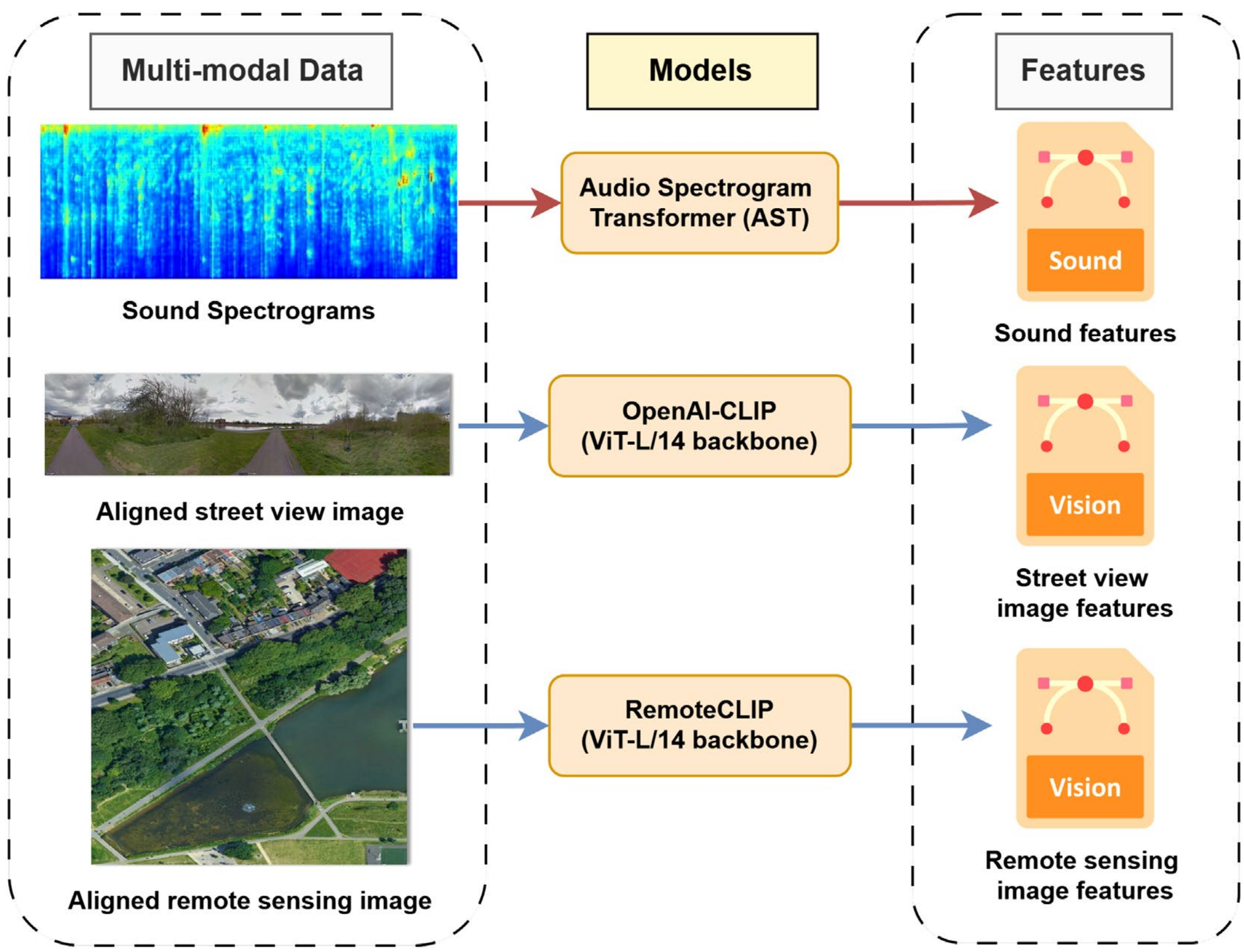

**FIGURE 3** | Modality-level summary of the feature extraction pipeline. Each input type (sound, street view, aerial image) is processed by a pre-trained model to produce corresponding feature embeddings.

Figure 3 presents a high-level overview of the feature extraction process. In this stage, sound spectrograms, street view imagery, and aerial images are processed using pretrained models: AST, OpenAI CLIP, and RemoteCLIP, respectively, to produce modality-specific embeddings. These embeddings constitute the input for subsequent similarity analysis tasks.

## 4.1 | Sound Feature Extraction

To represent the acoustic characteristics of urban environments, we employ the Audio Spectrogram Transformer (AST), a pre-trained transformer-based model designed for sound classification tasks. This study uses AST as the sound feature extractor; its output embeddings are treated as high-level representations of the input soundscape.

We select AST over alternatives such as PANNs (Kong et al. 2020) and PSLA (Gong et al. 2021b) because of its strong benchmark performance and architectural suitability for cross-modal embedding. Unlike convolutional networks that rely on local receptive fields, AST adopts a Vision Transformer backbone capable of modeling long-range temporal–spectral dependencies through self-attention. This design enables it to capture complex contextual structures—such as overlapping anthropogenic and biophonic signals—that frequently occur in urban environments. Benchmark evaluations on the large-scale AudioSet dataset demonstrate AST's superior accuracy (mean average precision = 0.485) compared with PANNs (0.439) and PSLA (0.474) under similar input resolution and parameter scales (Gong et al. 2021a). These results confirm that AST achieves stronger generalization for unstructured environmental audio while remaining compatible with publicly available pretrained weights. Such properties make AST particularly well aligned with our goal of computing semantic similarity across modalities in a shared latent space.

Each raw audio clip, represented in waveform format $x(t) \in \mathbf{R}^T$, is first transformed into a log-Mel spectrogram. This transformation involves applying the Short-Time Fourier Transform (STFT), followed by a Mel-scale filter bank and logarithmic compression:

$$S(f, \tau) = \log\left(\mathrm{Mel}\left(|\mathrm{STFT}(x(t))|^2\right)\right), \tag{1}$$

where $f$ denotes the Mel-frequency bin and $\tau$ represents the temporal frame index. The resulting spectrogram $S \in \mathbb{R}^{F \times T'}$ encodes time–frequency energy distribution and is treated as a 2D image-like input.

The spectrogram is divided into non-overlapping patches $\{P_1, P_2, \ldots, P_N\}$, each of which is flattened and linearly projected into an embedding space:

$$\mathbf{e}_i = \mathbf{W}_e \cdot \text{vec}(P_i) + \mathbf{b}_e, \quad (2)$$

where $\mathbf{e}_i \in \mathbb{R}^d$ is the embedding of patch $P_i$, and $\mathbf{W}_e$ and $\mathbf{b}_e$ are learnable parameters.

These token embeddings are then passed through a Vision Transformer (ViT) encoder, which models temporal and spectral dependencies using self-attention. The final output is a 768-dimensional feature vector that serves as a compact summary of the soundscape, encoding both spectral content (e.g., pitch, timbre) and temporal patterns (e.g., rhythm, modulation). We use this representation in downstream similarity computation, treating it as the sound modality's latent feature vector.

## 4.2 | Visual Feature Extraction

To extract semantic representations from visual data, we employ CLIP-based models due to their ability to map visual and textual modalities into a shared semantic space. This property enables direct comparison between images and audio embeddings, making CLIP well-suited for cross-modal alignment tasks.

CLIP (Contrastive Language–Image Pretraining) takes an image $I$ and a corresponding text caption $T$ as input, and encodes them into feature vectors using separate encoders:

$$\mathbf{z}_{\text{img}} = f_\theta(I), \quad \mathbf{z}_{\text{text}} = g_\phi(T) \quad (3)$$

where $f_\theta$ is a vision transformer (ViT) and $g_\phi$ is a text transformer. The model is trained to maximize the cosine similarity of matching pairs $(\mathbf{z}_{\text{img}}, \mathbf{z}_{\text{text}})$ while minimizing similarity to non-matching ones. After training, the image encoder $f_\theta$ is used in our work to extract visual features.

For street view imagery, we use OpenAI's pretrained CLIP model (ViT-L/14) to generate a 768-dimensional feature vector $\mathbf{z}_{\text{CLIP}}^{\text{street}}$. For aerial imagery, we adopt RemoteCLIP, which is fine-tuned on aerial image datasets to better capture overhead spatial structures. It produces a corresponding vector $\mathbf{z}_{\text{CLIP}}^{\text{aerial}} \in \mathbb{R}^{768}$.

All CLIP-based models in this study were used in their pretrained form without additional fine-tuning. The dataset size (175 locations across three cities) is relatively small, and fine-tuning large-scale vision–language models under such conditions may lead to overfitting or diminish the broad semantic alignment learned from large-scale image–text corpora. Our goal was therefore to evaluate the baseline zero-shot performance of pretrained CLIP and RemoteCLIP embeddings in urban contexts rather than to train new task-specific variants. While this approach may limit domain adaptation, it ensures model stability and comparability given the available data. Future work could explore lightweight adaptation methods such as Low-Rank Adaptation (LoRA) or prompt tuning once larger and more balanced datasets become available.

## 4.3 | Semantic Segmentation

To complement the embedding-based analysis, we also apply semantic segmentation to obtain interpretable, class-level representations of visual scenes. Segmentation provides explicit spatial and categorical information about objects and land cover types, enabling us to evaluate multimodal alignment in a more structured, human-interpretable form (Figure 4).

For street-level imagery, we use CLIPSeg, a prompt-based segmentation model derived from CLIP. Given an input image and a textual prompt (e.g., "building", "tree"), CLIPSeg predicts a dense binary mask highlighting the region associated with the prompt. It leverages the CLIP image encoder to preserve

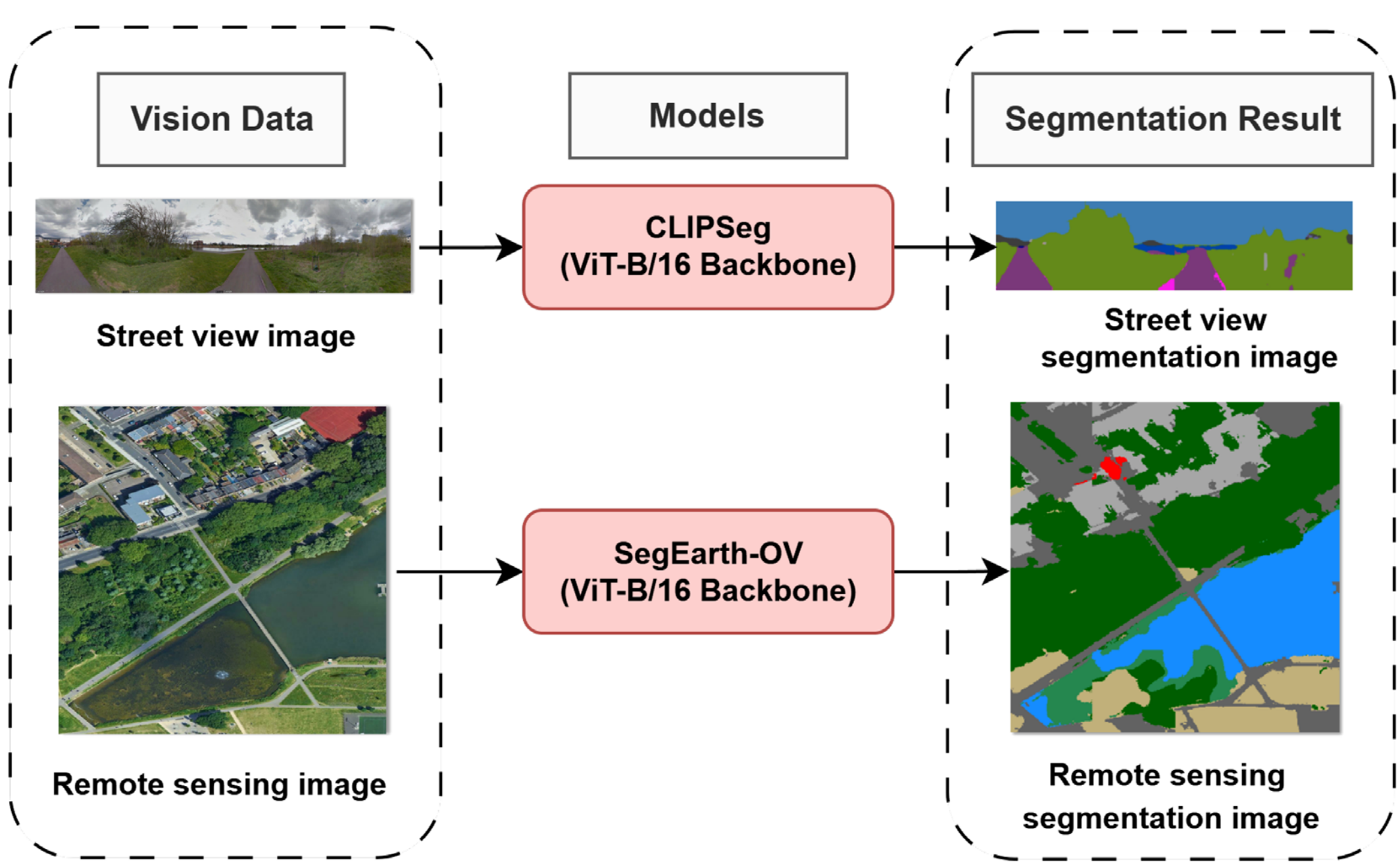

**FIGURE 4** | Semantic segmentation examples from street view (CLIPSeg; Lüddecke and Ecker 2022) and aerial imagery (Seg-Earth OV; Li, Liu, et al. 2024).

visual semantics and generalize to unseen categories without task-specific training.

For aerial imagery, we adopt Seg-Earth OV, a training-free open-vocabulary segmentation model also based on CLIP. Designed specifically for remote sensing, it enhances CLIP's patch-level features using an upsampling module (SimFeatUp) and applies a debiasing operation to remove global [CLS] token influence. This results in fine-grained and interpretable land cover maps across urban scenes.

Both segmentation outputs are post-processed to compute the proportion of pixels assigned to each class, forming a scene-level class distribution vector.

### 4.4 | Similarity Computation

To evaluate the alignment between soundscapes and visual scenes, we compute cross-modal similarity using two complementary approaches: embedding-based similarity and segmentation-based similarity.

In the embedding-based approach, we calculate cosine similarity between AST audio embeddings $\mathbf{z}_{\text{AST}}$ and visual embeddings derived from CLIP or RemoteCLIP:

$$\text{sim}_{\text{embed}} = \cos\left(\mathbf{z}_{\text{AST}}, \mathbf{z}_{\text{CLIP}}\right) \tag{4}$$

This measures the semantic closeness of sound and image representations in a shared latent space, capturing abstract features such as texture, context, and content.

In the segmentation-based approach, we compare scene-level class distribution vectors $\mathbf{p}_{\text{seg}} \in \mathbb{R}^{C}$, where $C$ is the number of semantic classes. These vectors represent the proportion of each visual class within the segmented image. Similarity is computed using cosine distance between paired sound-image samples:

$$\text{sim}_{\text{seg}} = \cos\left(\mathbf{p}^{\text{a}}_{\text{seg}}, \mathbf{p}^{\text{v}}_{\text{seg}}\right) \tag{5}$$

where $\mathbf{p}^{\text{a}}$ corresponds to the audio-mapped ecological distribution (via BGA) and $\mathbf{p}^{\text{v}}$ to the visual segmentation-derived distribution.

Together, these two similarity measures allow us to evaluate both abstract cross-modal alignment (through embeddings) and interpretable ecological correspondence (through segmentation). This comparative design constitutes a methodological contribution by establishing a consistent analytical protocol through which diverse sound–vision representations can be quantitatively evaluated and contrasted.

### 4.5 | Ecological Class Mapping: BGA

To interpret segmented scenes in ecological terms, we adopt the Biophony–Geophony–Anthrophony (BGA) classification framework (Pijanowski et al. 2011). This taxonomy enables us to relate visual land cover types to broad categories of environmental sound sources, enhancing the ecological interpretability of our cross-modal analysis (Rey-Baquero et al. 2021).

1. Biophony refers to sounds generated by non-human biological organisms, such as birds, insects, and other animals.
2. Geophony encompasses natural non-biological sounds, including wind, rain, flowing water, and other geophysical processes.
3. Anthrophony includes all human-related sound sources, such as traffic, construction, industrial machinery, and speech.

Visual segmentation outputs are mapped to these acoustic categories based on a predefined correspondence between land cover classes (e.g., vegetation, water, road, building) and their most likely sound-producing agents. For instance, green space classes are linked to Biophony, while impervious surfaces such as roads and buildings are linked to Anthrophony. This mapping allows us to estimate the likely ecological composition of soundscapes based solely on visual scene structure, offering a novel lens for analyzing sound–image alignment in urban environments.

The class-to-category values in Table 1 represent heuristic estimates derived from common ecological associations between land cover types and their most likely acoustic contributors, informed by prior work in soundscape ecology (Rey-Baquero et al. 2021). Rather than relying on training data or empirical measurements, we use these values as interpretable proxies to bridge visual semantics with ecological sound categories. This approach emphasizes conceptual clarity and domain-informed reasoning over strict numerical precision.

## 5 | Results

In this section, we present our findings from cross-modal similarity analysis across both embedding-based and segmentation-based representations. Results are grouped by feature type and evaluated for statistical significance and ecological interpretability.

### 5.1 | Embedding-Based Similarity

We computed cosine similarities between AST audio embeddings and visual embeddings extracted using CLIP (for street view) and RemoteCLIP (for aerial imagery), both employing the ViT-L/14 backbone. All correlation values were statistically significant ($p < 0.001$), indicating consistent cross-modal associations between visual and acoustic representations (Figure 5). Street-view imagery exhibited a stronger correlation with sound features ($r = 0.19$) than aerial imagery ($r = 0.14$), and combining both modalities further improved the relationship ($r = 0.21$), suggesting that ground-level and overhead perspectives encode complementary visual cues relevant to urban soundscapes. We computed cosine similarities between AST audio embeddings and visual embeddings extracted using CLIP (for street view) and RemoteCLIP (for aerial imagery), both employing the ViT-L/14 backbone. All correlation values

**TABLE 1** | Heuristic ecological soundscape mapping of segmentation classes.

| Class | Bio | Geo | Anthro | Notes |
|---|---|---|---|---|
| Aerial imagery | | | | |
| Grassland | 1.0 | 0.3 | — | Birds, insects |
| Forest/ Vegetation | 1.0 | 0.3 | — | High wildlife activity |
| Wetlands | 1.0 | 0.3 | — | Amphibians, aquatic insects |
| Waterbody | 0.3 | 1.0 | — | Water flow, aquatic birds |
| Bare Land | 0.1 | 0.1 | 1.0 | Sparse habitat, occasional use |
| Road/ Sidewalk | 0.1 | — | 1.0 | Urban traffic |
| Building | 0.1 | — | 1.0 | Mechanical/ human noise |
| Vehicles | — | — | 1.0 | Engine sounds |
| Cropland | 1.0 | — | 0.3 | Farming activity |
| Street view imagery | | | | |
| Road | — | — | 1.0 | Traffic noise |
| Sidewalk | 0.3 | — | 1.0 | Pedestrians, ambient urban sound |
| Building | 0.3 | — | 1.0 | Urban structures |
| Vegetation | 1.0 | 0.3 | — | Birds, insects |
| Waterbody | 1.0 | 1.0 | — | Flowing water |
| Person | — | — | 1.0 | Human voice, footsteps |
| Car, Truck, Bus, etc. | — | — | 1.0 | Engine noise |

*Note:* Values range from 0 (none) to 1.0 (strong); "–" denotes no expected contribution.
Abbreviations: Anthro, Anthrophony; Bio, Biophony; Geo, Geophony.

were statistically significant ($p < 0.001$), indicating consistent cross-modal associations between visual and acoustic representations (Figure 5). Street-view imagery exhibited a stronger correlation with sound features ($r = 0.19$) than aerial imagery ($r = 0.14$), and combining both modalities further improved the relationship ($r = 0.21$), suggesting that ground-level and overhead perspectives encode complementary visual cues relevant to urban soundscapes.

Although these coefficients are low in magnitude, such results are theoretically expected for heterogeneous modalities. Cosine correlation between audio and image embeddings is intrinsically bounded by the limited mutual information shared between them: visual data capture static geometry and material appearance, whereas acoustic data reflect dynamic and often occluded processes such as motion, human activity, and mechanical resonance. Consequently, the embedding spaces form distinct yet intersecting manifolds rather than identical feature distributions. Even purpose-built audiovisual correspondence models achieve only moderate alignment—for example, $L^3$-Net reports an average cross-modal retrieval score of nDCG @ 30 = 0.385 (audio → image) on the AudioSet-Instruments dataset (Arandjelović and Zisserman 2018)—demonstrating the inherent information bottleneck between sight and sound. Similarly, large-scale multimodal frameworks such as ImageBind (Girdhar et al. 2023) achieve moderate zero-shot retrieval accuracy between audio and image modalities (Top-1 ≈ 30%–36%), substantially lower than for text–image pairs, confirming that even unified embeddings capture only partial cross-modal dependence. In our case, this limitation is further amplified by the use of pretrained but independently optimized models (AST for audio and CLIP/RemoteCLIP for imagery) without joint fine-tuning.

Across the three cities, consistent patterns emerge (Figure 6). Street-view embeddings correlate more strongly with sound features than aerial embeddings in London ($r \approx 0.20$ vs. 0.14) and Tokyo ($r \approx 0.26$ vs. 0.23). New York, however, diverges from this trend, with aerial similarity showing minimal correlation ($r \approx 0.04$) due to extreme verticality and occlusion, where top-down imagery captures roof surfaces but not sound-relevant street-level elements. The combined-view correlation improves slightly ($r = 0.21$ overall), reaffirming the complementary nature of multi-perspective visual information.

These trends are further illustrated in Figures 7 and 8, which visualize representative cases of alignment and divergence. In aerial views (Figure 7), high visual similarity does not always yield similar soundscapes because key acoustic sources—such as biophony or anthropogenic noise—may be hidden from overhead perspectives. Conversely, visually distinct but ecologically similar areas (e.g., vegetated or waterfront zones) can produce analogous soundscapes. In street-level scenes (Figure 8), high visual similarity may mask acoustic variation driven by local traffic density or transient human activities, whereas acoustically similar environments sometimes arise from semantically distinct but functionally similar elements (e.g., trees, road traffic).

In summary, while the correlation between visual and acoustic representations is modest in absolute value, it is statistically robust and theoretically consistent with the limited mutual information shared between sound and vision. Multi-perspective imagery provides complementary cues that together yield a more comprehensive depiction of the urban acoustic environment. These findings directly address RQ1, confirming that environmental soundscapes exhibit measurable but bounded correspondence with visual representations of urban form. The results demonstrate that sound–vision alignment is stronger for street-level imagery than for aerial views, owing to the former's richer capture of proximal and human-scale acoustic cues. Furthermore, the improvement observed when combining both perspectives underscores that multi-view visual information can enhance the perceptual completeness of cross-modal representations. Together, these results establish a quantitative baseline for understanding how environmental sounds relate to visual urban morphology across spatial perspectives.

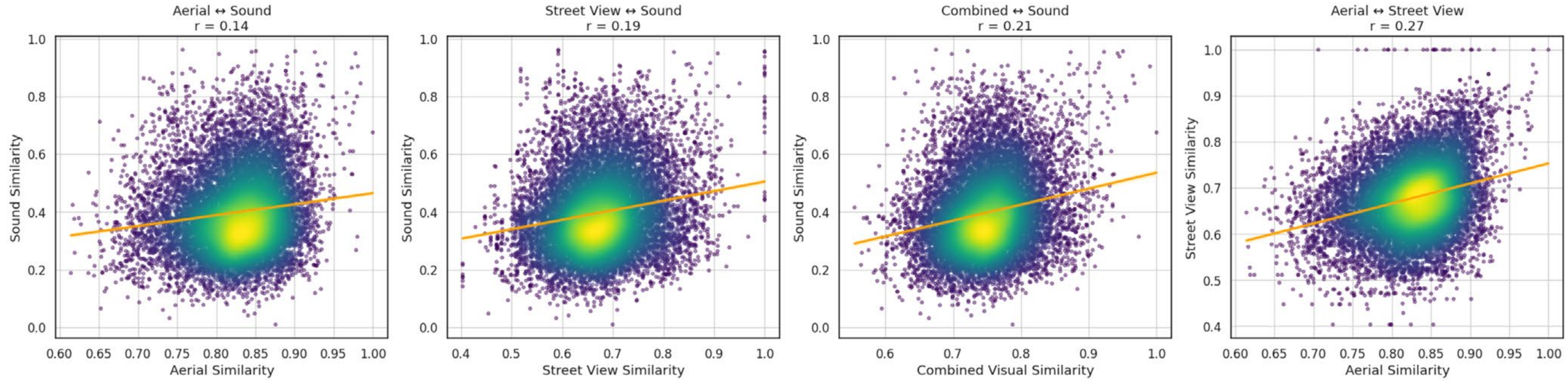

**FIGURE 5** | Correlation between audio and visual embeddings. Street-view embeddings exhibit stronger correlation with sound features. Combining both street and aerial views yields the highest performance, supporting the complementarity of these perspectives.

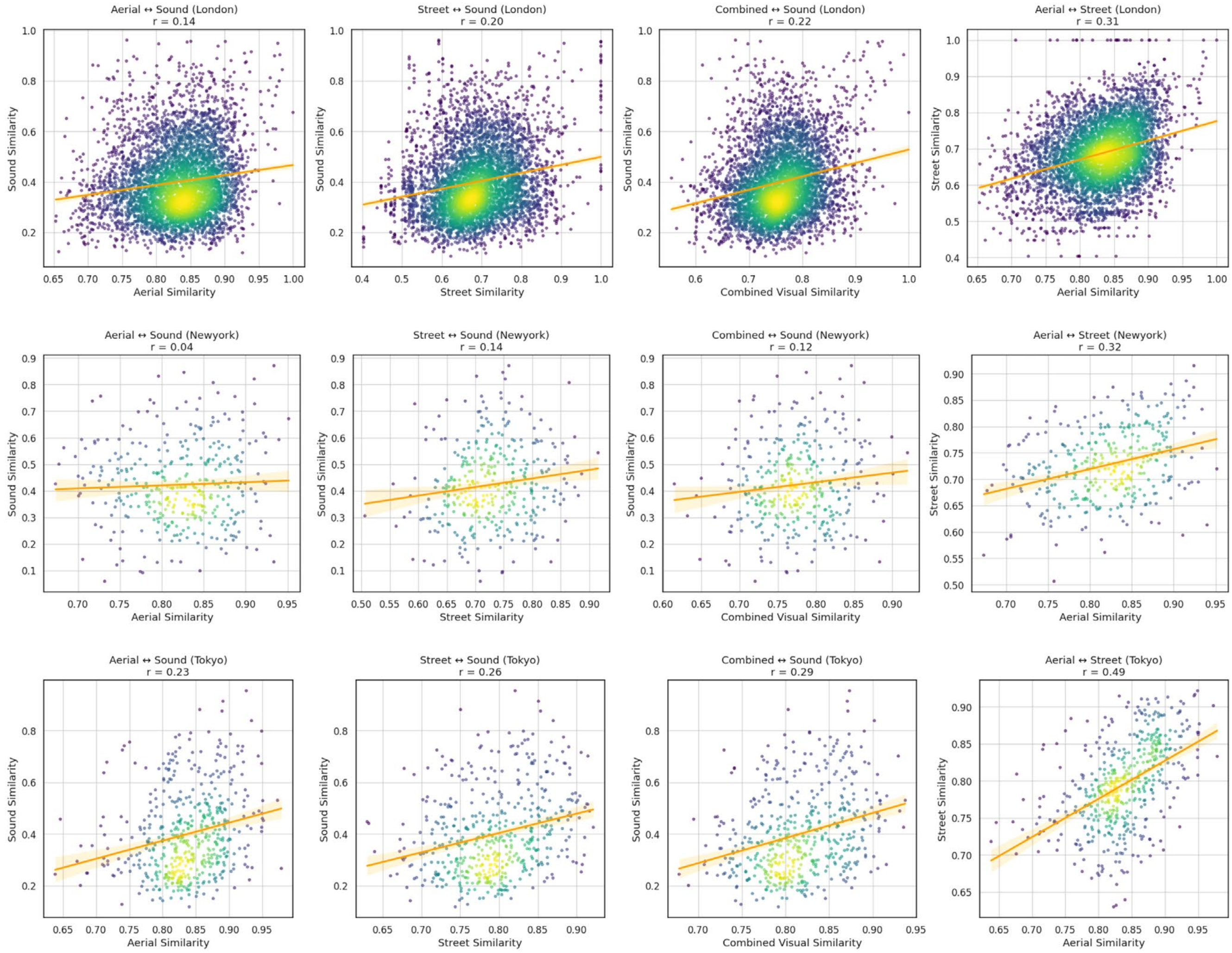

**FIGURE 6** | City-level correlation between sound similarity and visual similarity across modalities. Street-view and combined embeddings generally outperform aerial imagery alone, particularly in dense urban areas like New York City.

## 5.2 | Segmentation-Based Similarity

Semantic segmentation maps were converted into class-wise pixel distribution vectors and compared with sound embeddings using cosine similarity. The analysis reveals a clear contrast between aerial and street-view modalities (Figure 9). Across all samples, aerial segmentation similarity shows a statistically significant positive correlation with sound similarity ($r = 0.16, p = 1.9 \times 10^{-48}$), whereas street-view segmentation exhibits only a weak relationship ($r = 0.03, p = 2.1 \times 10^{-3}$). The difference of 0.13 confirms that large-scale land-cover composition visible from overhead perspectives provides stronger and more consistent cues to acoustic variation than ground-level segmentation.

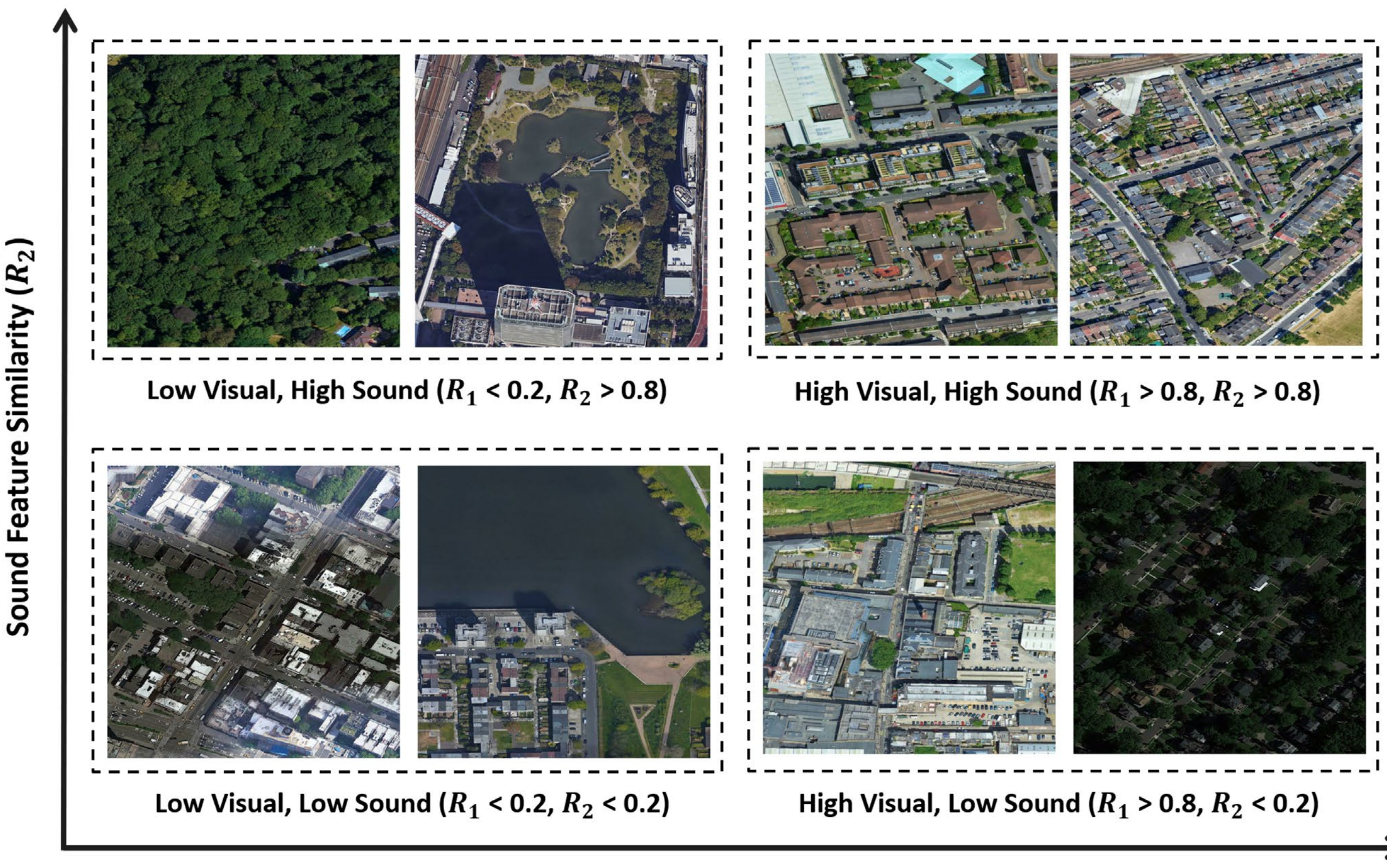

**FIGURE 7** | Visual examples from aerial imagery illustrating visual–acoustic alignment and divergence. High visual similarity does not always yield similar soundscapes due to the limited visibility of sound sources, while visually distinct but ecologically similar areas may share acoustic characteristics.

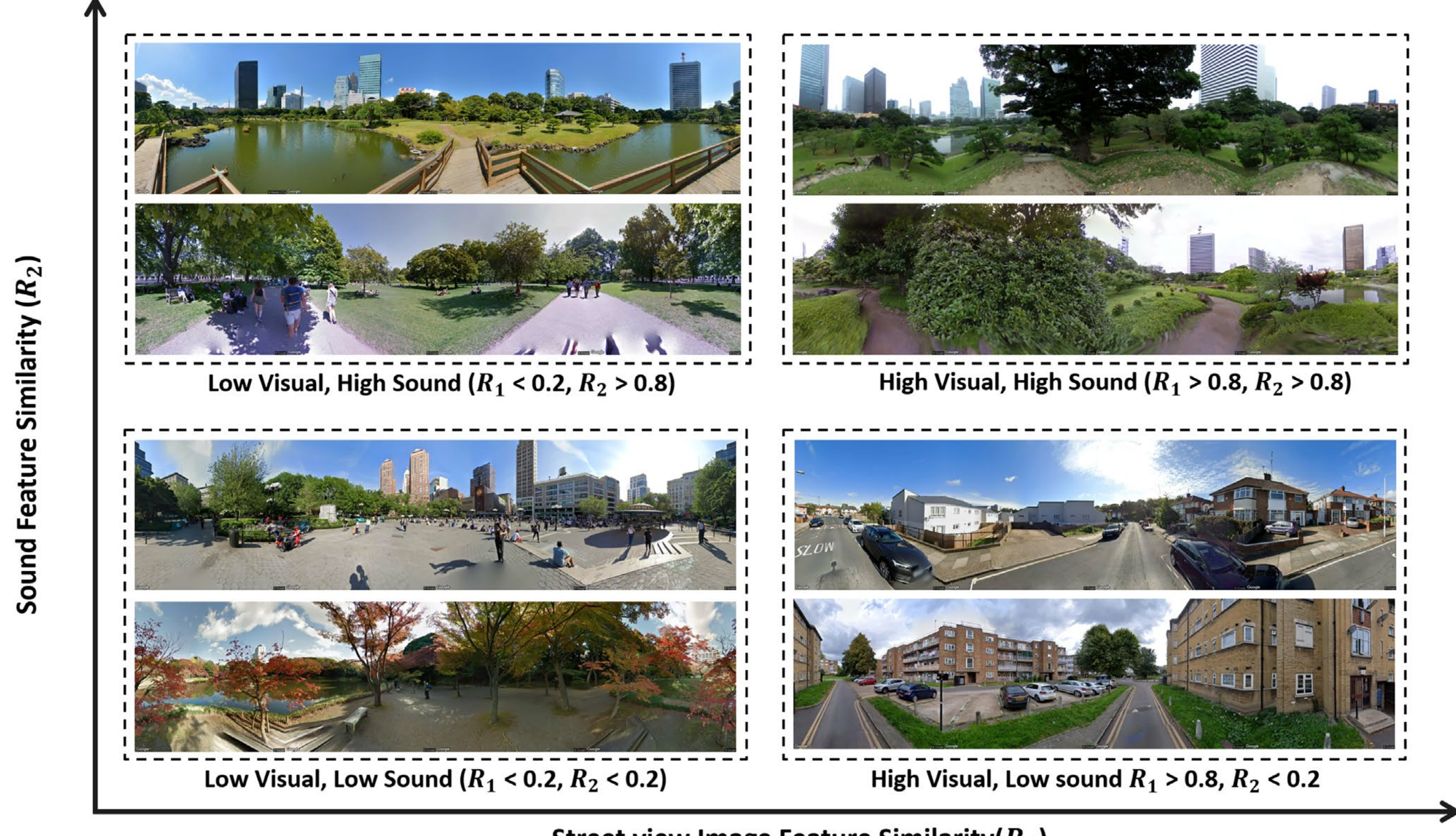

**FIGURE 8** | Examples from street-view imagery showing varying degrees of visual–acoustic alignment. Localized activities, vegetation, and traffic contribute to differences in soundscapes even among visually similar scenes.

City-level patterns follow the same hierarchy. In Tokyo, aerial segmentation yields the highest correlation with sound similarity ($r = 0.27, p = 1.0 \times 10^{-9}$), consistent with the abundance of vegetated and park areas. New York also shows a positive relationship ($r = 0.18, p = 5.0 \times 10^{-4}$), while London presents a weaker correlation ($r = 0.12, p = 1.1 \times 10^{-17}$). Street-view segmentation remains near zero in all cases ($r = -0.03$ in New York, $r = 0.05$ in London, $r = 0.08$ in Tokyo), indicating that current ground-level semantic segmentation fails to capture visual structures that meaningfully correspond to urban soundscapes (Figure 10).

This outcome contrasts with the embedding-based results, where street-view imagery correlated more strongly with sound than aerial views. The inversion reflects a fundamental distinction between the two modeling paradigms. Embedding-based similarity emphasizes perceptual and contextual semantics captured by pretrained vision–language models, which are more representative of the sensory experience at street level. In contrast, segmentation-based similarity isolates structural and compositional information, land cover, vegetation, and built intensity, which are better captured from the overhead perspective. Hence, aerial segmentation excels at identifying the physical context that shapes acoustic environments, while street-view embeddings are more sensitive to perceptual cues of the soundscape.

### 5.3 | Ecological Alignment via BGA Categories

To examine the ecological interpretability of visual–acoustic correspondence, we applied the predefined BGA mapping (Table 1) to group segmented classes into Biophony, Geophony, and Anthrophony categories. For each modality, category-wise similarity was computed between the BGA vectors and AST-based sound embeddings (Figure 11).

Across the entire dataset, aerial imagery demonstrates consistent positive alignment with acoustic similarity for all ecological categories. Anthrophony yields the highest correlation ($r = 0.145,\ p = 2.7 \times 10^{-59}$), followed by Biophony ($r = 0.130,\ p = 6.1 \times 10^{-48}$) and Geophony ($r = 0.070,\ p = 9.0 \times 10^{-15}$). In contrast, correlations derived from street-view segmentation are negligible: Biophony ($r = -0.023,\ p = 1.1 \times 10^{-2}$), Anthrophony ($r = -0.005,\ p = 0.55$), and Geophony ($r = 0.024,\ p = 6.4 \times 10^{-3}$). The mean difference between aerial and street-view correlations across categories is approximately 0.10, confirming that overhead land-cover composition provides more stable ecological cues to urban soundscapes than local street-level visual patterns.

City-level results (Figures 12 and 13) reinforce these global trends while revealing contextual nuances. In aerial imagery, Tokyo shows the strongest ecological alignment, particularly for Biophony ($r = 0.23$) and Anthrophony ($r = 0.25$), reflecting the city's large proportion of parks and vegetated belts. London displays moderate but significant correlations across all categories, whereas New York's alignment is generally weak, consistent with its highly built form and limited ecological heterogeneity.

For street-view imagery, most correlations remain close to zero. London and New York show minimal alignment across categories, with Anthrophony in New York ($r = 0.14,\ p < 0.01$) as a minor exception. Tokyo again presents some signal, where Biophony ($r = -0.14$) and Anthrophony ($r = 0.26$) are both significant but opposite in direction, indicating complex local variation in how visible vegetation and human activity relate to perceived sound.

Beyond correlation magnitudes, spatial interpretation provides additional ecological insight. Examining the geographic distribution of similarity scores reveals that higher vision–sound alignment tends to occur in vegetated, waterfront, or low-traffic zones, where both visual and acoustic environments are ecologically coherent. Dense commercial and high-rise districts, by contrast, display weaker or inconsistent coupling, reflecting the masking effects of built form and transient anthropogenic activity. These patterns suggest that cross-modal similarity captures not only semantic correspondence but also underlying spatial organization within the urban landscape, linking acoustic dynamics to the morphology of green, mixed, and dense urban typologies.

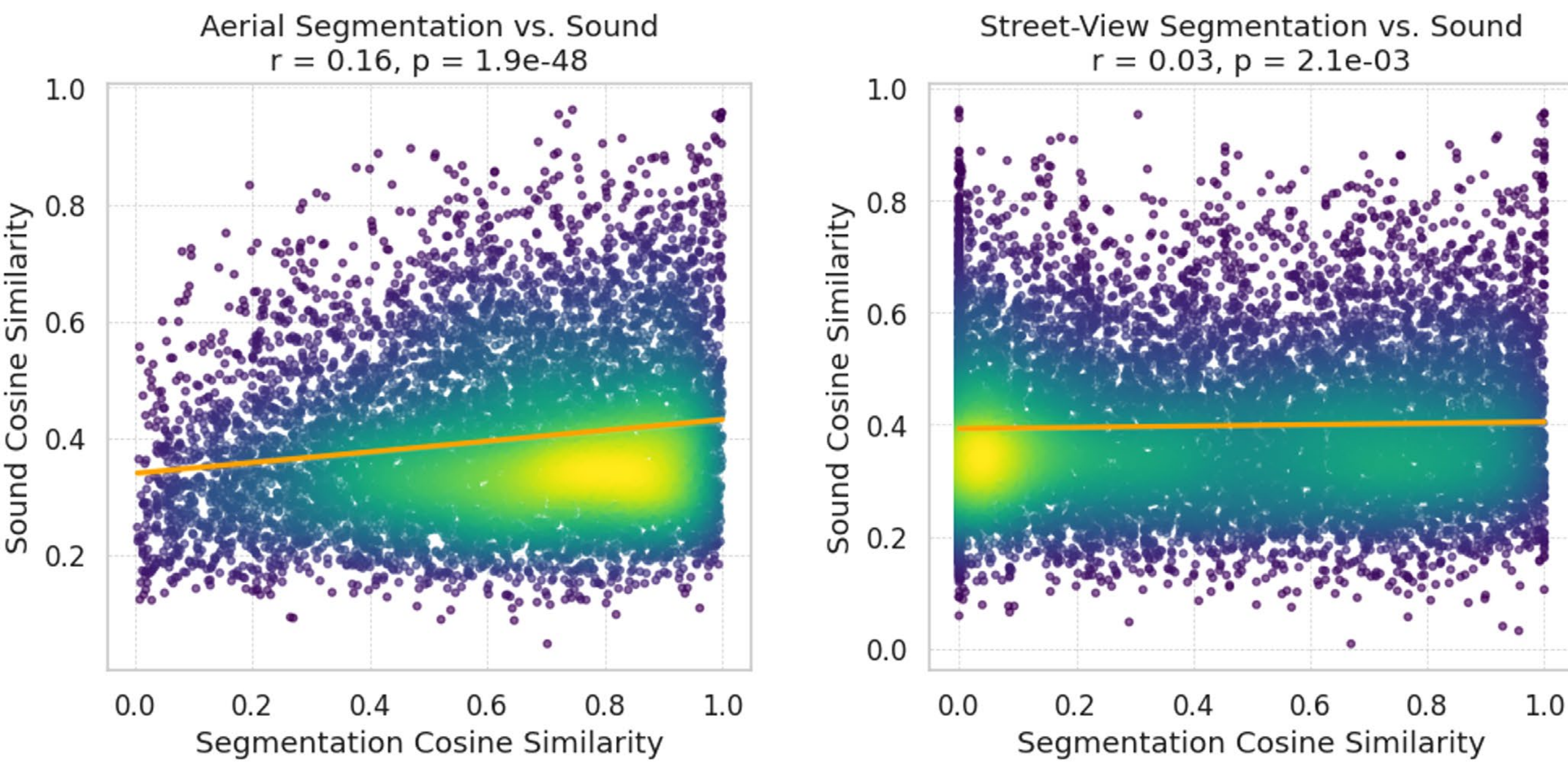

**FIGURE 9** | Segmentation-based similarity across all samples. Aerial segmentation shows a significant positive correlation with sound similarity, while street-view segmentation remains weak.

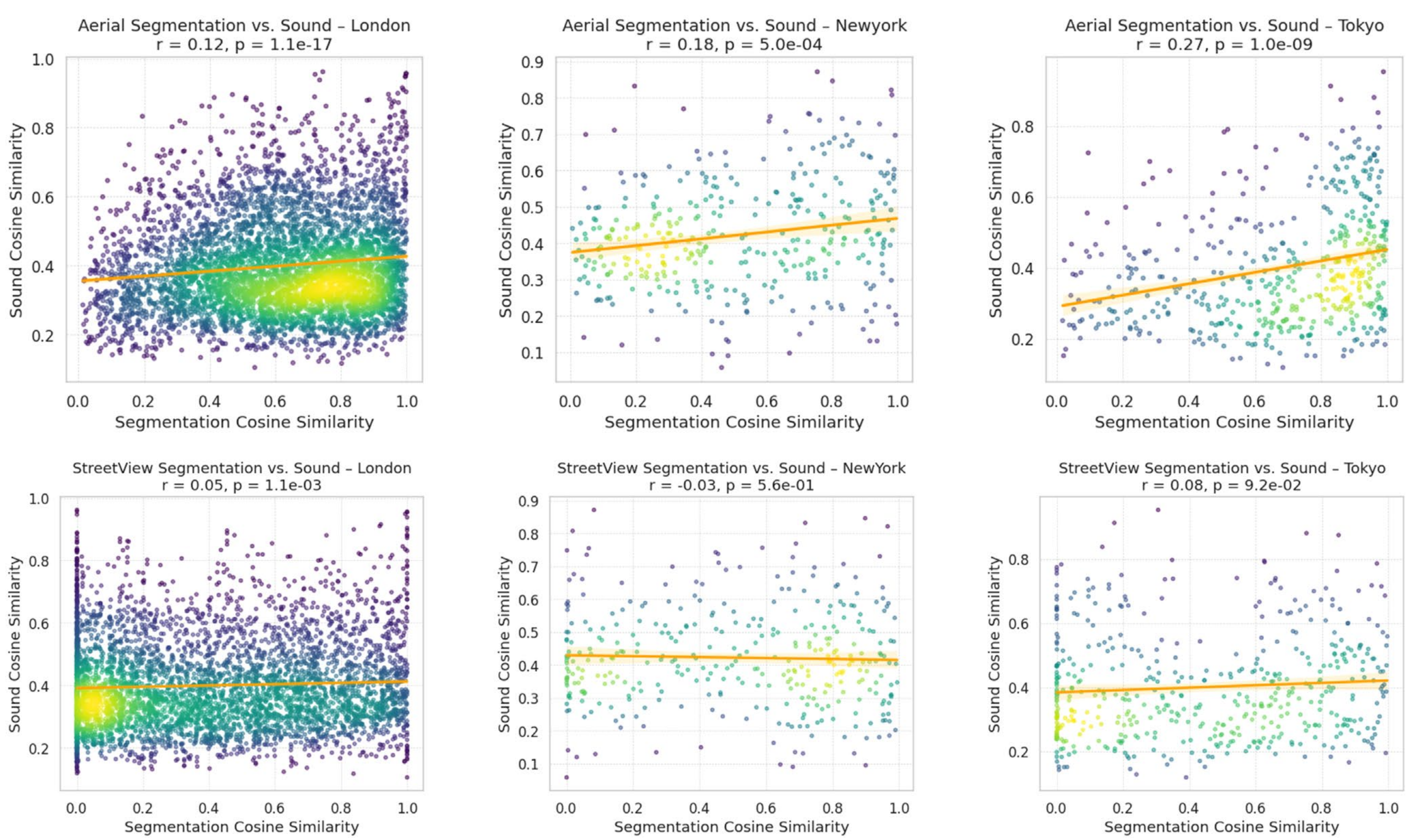

**FIGURE 10** | City-specific correlation between segmentation-based similarity and sound similarity. Aerial segmentation consistently outperforms street-view segmentation, especially in vegetation-rich areas such as Tokyo.

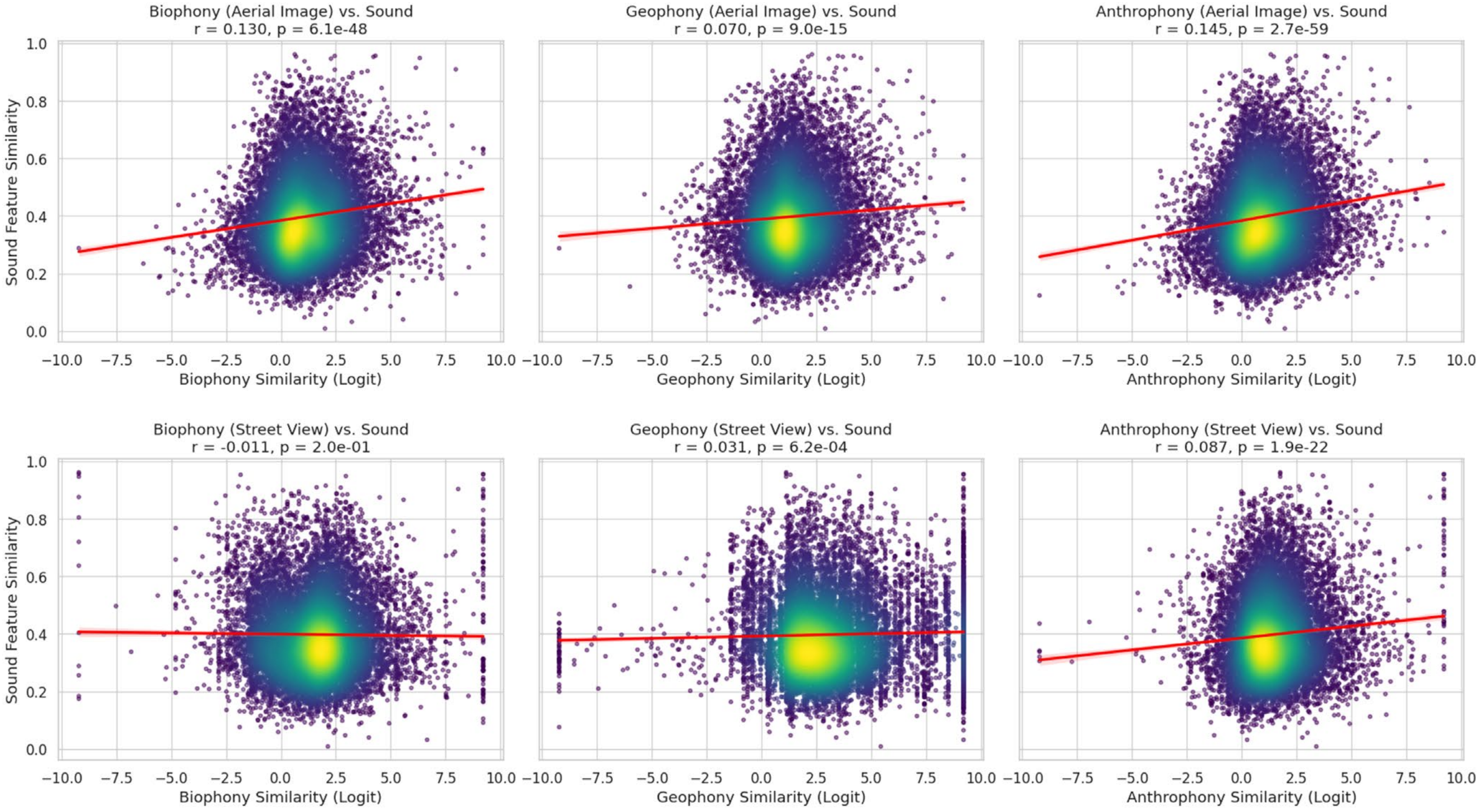

**FIGURE 11** | Correlation between AST-based sound similarity and BGA category similarity derived from aerial and street-view segmentation. Aerial imagery exhibits clearer and stronger ecological alignment than street-view imagery across all three categories.

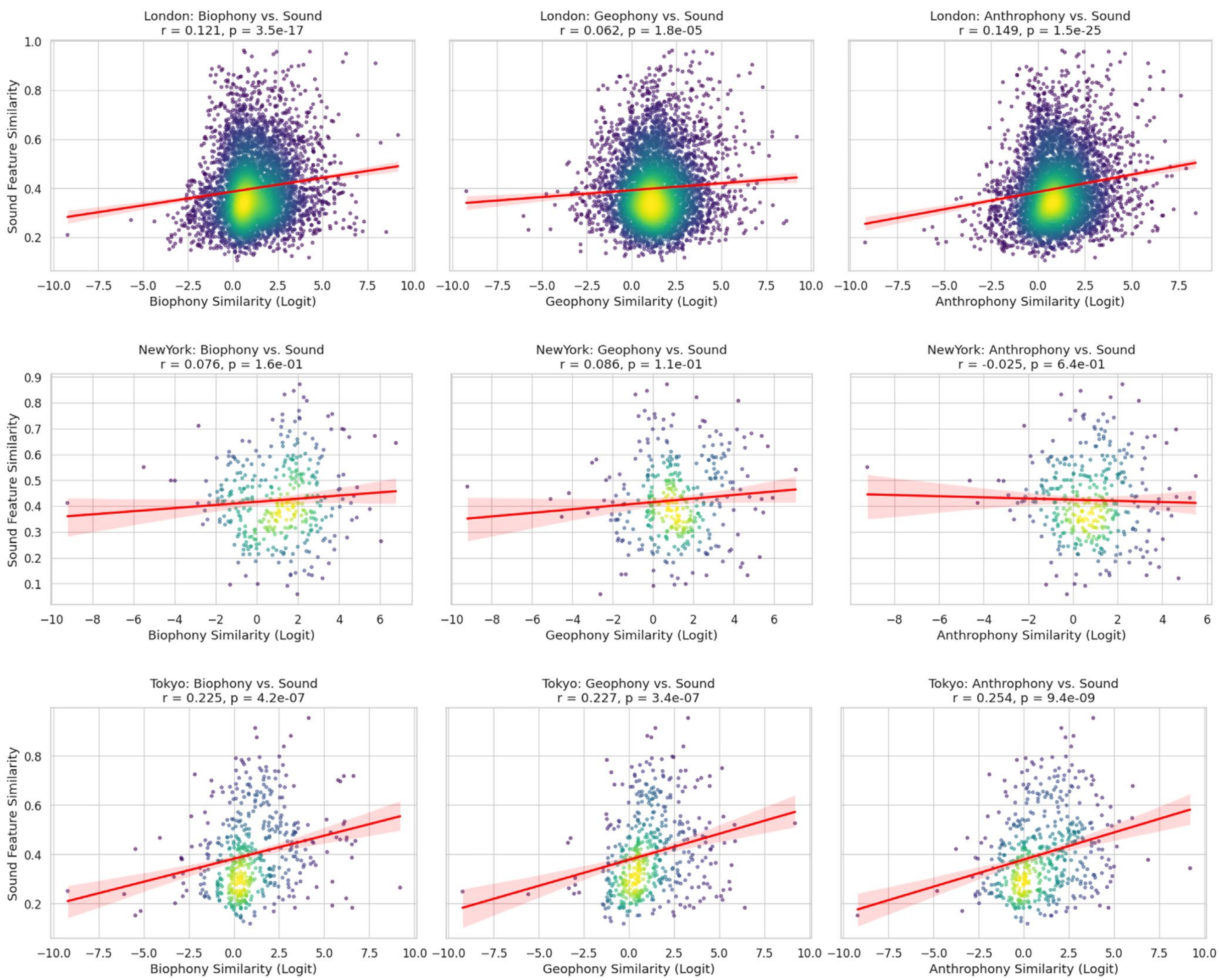

**FIGURE 12** | City-level correlation between aerial BGA similarity and sound similarity. Tokyo exhibits the strongest alignment across ecological categories, followed by London.

Taken together, these results demonstrate that ecological alignment between vision and sound is substantially stronger from the aerial perspective. The aerial viewpoint captures the spatial configuration of vegetation, water, and built surfaces that directly shape environmental acoustics, while street-level imagery reflects transient perceptual details that vary by time and activity. In other words, segmentation-based BGA metrics emphasize the structural context of sound generation rather than its perceptual experience.

These findings address RQ2: segmentation-based approaches, especially when interpreted through ecological categories such as BGA, provide transparent yet coarse approximations of acoustic similarity. Although less semantically rich than embedding-based models, they yield interpretable ecological indicators useful for large-scale environmental assessments. In contrast, street-view segmentation contributes inconsistently, underscoring the importance of vantage point and representation strategy in multimodal urban analysis.

## 6 | Discussion

### 6.1 | Asymmetry in Modality Performance: Street View vs. Aerial Imagery

A prominent finding, as detailed in Section 5.1, is the consistent difference in how street-view and aerial imagery performed across the two analytical methods. Embeddings from street-view images showed a stronger and more direct alignment with auditory features when compared directly, with a correlation of $r = 0.19$ with sound features, which was higher than aerial imagery's correlation of $r = 0.14$. Combined, these visual modalities yielded an even higher correlation of $r = 0.21$, suggesting their complementary nature. In contrast, the semantic segmentation of aerial imagery, as discussed in Sections 5.2 and 5.3, proved more effective and offered more interpretable ecological categorizations when applying the Biophony-Geophony-Anthrophony (BGA) framework. For instance, aerial segmentation similarity showed a statistically significant positive correlation with sound similarity ($r = 0.14$), while street view segmentation had a much weaker relationship ($r = 0.05$). This divergence implies that each type of visual data captures unique, yet complementary, aspects of the urban soundscape. Street-level imagery, by its very nature, provides an up-close perspective of the urban setting. This proximity makes it more adept at capturing fine-grained contextual details, such as the immediate presence of vehicles, pedestrian activity, or nearby vegetation, that are more directly and temporally linked to the sounds recorded in that specific location. Advanced embedding models, such as CLIP, appear proficient at encoding these subtle, nuanced details, effectively preserving them within the embedding space for comparison across different modalities. This is reflected in the city-specific results where street view embeddings generally outperformed aerial embeddings in correlating with sound, such as in London ($r = 0.20$ for street vs. $r = 0.14$ for aerial) and Tokyo ($r = 0.26$ for street vs. $r = 0.23$ for aerial). Conversely, aerial imagery offers a broader, bird's-eye view, emphasizing larger land cover types and their spatial arrangements. While this perspective might be less sensitive to fleeting, localized sounds, it aligns more closely with persistent ecological categories and general soundscape characteristics that can be inferred from landscape-level features. This inherent characteristic explains its superior performance in the BGA-based ecological mapping (Section 5.3), where overarching environmental classifications are of primary importance. For example, in aerial imagery, all three BGA categories (Anthrophony, Biophony, and Geophony) demonstrated statistically significant positive correlations with acoustic similarity, with Anthrophony showing the strongest performance ($r = 0.145$).

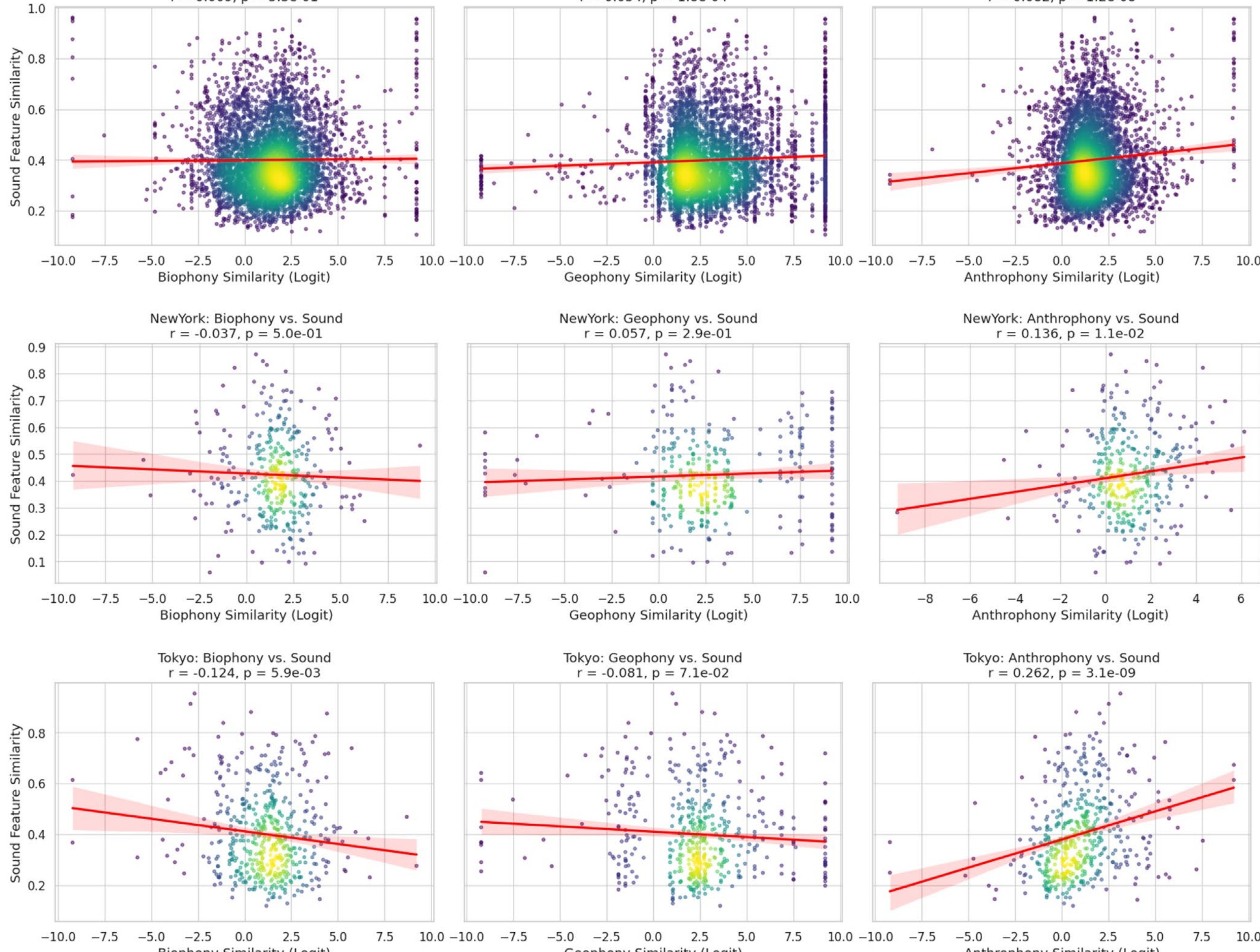

**FIGURE 13** | City-level correlation between street-view BGA similarity and sound similarity. Most correlations are weak or non-significant, with notable exceptions in Tokyo and New York.

The observed asymmetry between aerial and street-view modalities highlights an opportunity for multi-scale fusion rather than a limitation of either view. Future research could integrate the complementary advantages of both perspectives by developing hierarchical or cross-attention fusion frameworks that jointly encode macro-scale spatial context from aerial imagery and fine-grained perceptual cues from street-view imagery. Such architectures could align visual features across scales before linking them to sound representations, thereby improving both accuracy and interpretability of sound–vision alignment. Recent progress in vision–language models (VLMs) such as Qwen3-VL and similar multimodal transformers provides a practical foundation for this direction: these models can already process heterogeneous visual inputs and reason across modalities through unified embedding spaces. Extending such VLMs to incorporate dual-view visual streams with environmental sound input would enable more holistic and explainable modeling of urban acoustic environments.

### 6.2 | The Underperformance of Street-View Segmentation: A Closer Look

A noteworthy outcome, highlighted in Section 5.2, was the comparatively weak and often insignificant correlation observed between sound features and the similarity scores derived from the segmentation of street-view imagery. Across all samples, street view segmentation yielded a correlation of only $r = 0.05$ with sound similarity, and city-specific analyses showed this trend consistently, with correlations near zero or even slightly negative (e.g., $r = -0.03$ in New York, $r = 0.05$ in London, and $r = 0.08$ in Tokyo, all statistically weak or non-significant). This phenomenon can be traced to several contributing factors. Firstly, the inherent characteristics of the segmentation models used, like CLIPSeg, need to be considered. While these models are capable of identifying a diverse range of object categories, their output may sometimes lack the pixel-level precision or the consistent geometric alignment with complex urban scenes that is necessary for a nuanced interpretation of acoustic environments. The process of translating a visual object mask directly into an acoustic counterpart is not always straightforward or reliable. Secondly, urban soundscapes are inherently dynamic and are frequently shaped by temporary, ephemeral events such as a passing emergency vehicle, transient construction noise, or conversations. A static visual segmentation, regardless of its accuracy, fails to capture this crucial temporal dimension that defines many sound events. This is a key challenge when trying to link static visual data to dynamic auditory experiences. Furthermore, a critical disparity often exists between what is visually salient in an image and what has a significant auditory impact. For instance, a relatively small visual element, such as a distant but loud motorcycle or an inconspicuous air conditioning unit, might dominate the acoustic environment of a location. Yet, this sound source could occupy a minimal portion of the segmented image, or it might even be entirely hidden from view. This fundamental disconnect between visual prominence and acoustic significance likely underlies the limited effectiveness of street-level image segmentation in predicting soundscape characteristics, as evidenced by the weak BGA category correlations for street view, where Biophony and Anthrophony showed negligible or negative correlations (e.g., $r = -0.023$ for Biophony).

### 6.3 | Practical Implications and Potential Applications

Overall, this study advances the emerging field of multimodal urban sensing and contributes to urban science by revealing the multifaceted relationships between urban visual structures and acoustic environments. First, our exploration demonstrates the potential of using existing visual data to infer or monitor soundscape characteristics in urban areas lacking acoustic sensors. Furthermore, this study presents a methodological framework for incorporating sound as an information source into large-scale geospatial analytics. Finally, it highlights the importance of integrating soundscapes into urban planning, design, and landscape architecture, as sound is closely linked to visual elements and may exert a compounded influence on people's perception, mood, and well-being within multisensory environments.

### 6.4 | Limitations and Future Directions

While this study presents compelling findings, several inherent limitations should be acknowledged as directions for future research.

A primary methodological limitation lies in model generalization. The deep learning models used in this study (AST, CLIP, RemoteCLIP, CLIPSeg, and Seg-Earth OV) were applied in their pretrained form without task-specific fine-tuning on the compiled urban sound–image dataset. This approach ensured comparability and generalizability across modalities but restricted optimization for this particular task. The resulting cross-modal correlations, although statistically significant, remained moderate, with the highest combined visual–sound correlation reaching $r = 0.21$. Future research could explore lightweight domain-adaptation techniques such as Low-Rank Adaptation (LoRA) or prompt tuning to enhance urban-specific sensitivity while retaining the broad semantic consistency of pretrained models.

A second limitation concerns the assumption of spatio-environmental synchronicity, namely that the sound and the image data reflect the same location and environmental state. In complex urban topographies, acoustic propagation and occlusion effects can decouple the auditory and visual domains. Sounds often originate from sources beyond the immediate visual field or are affected by reflection and shielding from surrounding structures. This mismatch is most apparent in dense vertical cities such as New York, where aerial visual similarity showed minimal correlation with sound similarity ($r = 0.04$). Addressing this issue may require pairing ground-truth acoustic measurements with synchronized imagery or applying acoustic-simulation models to estimate how urban morphology shapes sound diffusion.

A third limitation relates to data quality and uncertainty in both modalities. Publicly sourced sound recordings from Radio Aporee and imagery from Google Earth and Street View may contain background noise, compression artifacts, or inconsistent lighting conditions. Although extensive filtering was performed, including the removal of recordings dominated by speech or indoor noise, residual bias and data noise likely contributed to the modest alignment values observed. Future work could incorporate noise-robust feature extraction and uncertainty-aware modeling, supported by data-augmentation strategies such as spectrogram mixing, random cropping, or photometric perturbation, together with explicit confidence estimation in embedding alignment.

Beyond these practical considerations, several methodological directions remain open. Developing unified cross-modal transformers or retrieval-based fusion models that jointly encode spatial and acoustic cues may better capture latent relationships between modalities. Integrating fine-grained street-view information with macro-scale aerial context through multi-scale fusion or hierarchical attention could further improve interpretability and alignment accuracy. Finally, evaluation frameworks should expand beyond correlation metrics to include retrieval-based measures (Top-$k$, nDCG) and higher-order dependence statistics such as the Hilbert–Schmidt Independence Criterion (HSIC) or canonical-correlation analysis (CCA), providing a more rigorous assessment of cross-modal correspondence.

Overall, these directions highlight both the promise and the challenges of multimodal environmental sensing. Future progress will depend on advances in dataset curation, model adaptation, and fusion architectures capable of bridging the perceptual and structural aspects of urban soundscapes.

## 7 | Conclusion

This study explored the alignment between environmental soundscapes and visual representations of urban environments using a multimodal framework that integrates street-level and aerial imagery with geo-referenced sound recordings. By comparing feature-based and segmentation-based approaches across modalities, we assessed the extent to which different visual representations encode meaningful acoustic information.

Our results show that embedding-based representations from CLIP and RemoteCLIP align more strongly with sound embeddings than semantic segmentation outputs. Street view imagery performs best in capturing fine-grained auditory semantics, while remote sensing segmentation—despite lower overall correlation—offers more interpretable insights through ecological class mapping. Specifically, the Biophony–Geophony–Anthrophony (BGA) classification reveals a structured relationship between land cover and likely sound types, especially in aerial views.

These findings demonstrate that visual features can serve as useful proxies for environmental sound, with potential applications in soundscape-aware urban planning, acoustic ecology, and large-scale multimodal sensing. The ability to infer ecological sound composition from visual data opens new pathways for passive, scalable acoustic monitoring in cities.

Future work may focus on improving alignment through fine-tuning of visual and acoustic models, incorporating temporal context, or developing unified cross-modal training objectives. Expanding the framework to include richer contextual data (e.g., time of day, weather, human activity) may also enhance robustness and ecological interpretability.

### Acknowledgments

Declaration of Generative AI and AI-Assisted Technologies in the Writing Process: During the preparation of this work the authors used Gemini 2.5 in order to correct grammar and improve clarity of the language. After using this AI tool, the authors reviewed and edited the content as needed and took full responsibility for the content of the published article.

### Conflicts of Interest

The authors declare no conflicts of interest.

### Data Availability Statement

The data that support the findings of this study are available in Radio Aporee at https://aporee.org/aporee.html.